*Article*

# PCB Component Detection using Computer Vision for Hardware Assurance

Wenwei Zhao[1,*], Suprith Gurudu[1], Shayan Taheri[1], Shajib Ghosh[1], Mukhil Azhagan Mallaiyan Sathiaseelan[1], Navid Asadizanjani[1]

1. Electrical and Computer Engineering Department, University of Florida
* Correspondence: wenwei.zhao@ufl.edu (W.Z.)

**Abstract:** Printed Circuit Board (PCB) assurance in the optical domain is a crucial field of study. Though there are many existing PCB assurance methods using image processing, computer vision (CV), and machine learning (ML), the PCB field is complex and increasingly evolving so new techniques are required to overcome the emerging problems. Existing ML-based methods outperform traditional CV methods, however they often require more data, have low explainability, and can be difficult to adapt when a new technology arises. To overcome these challenges, CV methods can be used in tandem with ML methods. In particular, human-interpretable CV algorithms such as those that extract color, shape, and texture features increase PCB assurance explainability. This allows for incorporation of prior knowledge, which effectively reduce the number of trainable ML parameters and thus, the amount of data needed to achieve high accuracy when training or retraining an ML model. Hence, this study explores the benefits and limitations of a variety of common computer vision-based features for the task of PCB component detection. Results of this study indicate that color features demonstrate promising performance for PCB component detection. The purpose of this paper is to facilitate collaboration between the hardware assurance, computer vision, and machine learning communities.

## 1. Introduction

Modern electronic systems ranging from personal computers and mobile devices to critical government, military, and medical infrastructures use Printed Circuit Boards (PCB) as functional blocks that connect different electrical components, traces, and vias [1]. With the advancement of semiconductor industries, the design of these PCBs is getting highly complex, having multiple layers with hidden vias and embedded passive components to meet the requirements of advanced systems. Incorporating such complexities creates a great opportunity for the potential attackers to maliciously modify the design and thus expose the PCB supply chain to vulnerabilities [2]. Moreover, due to the dominant trend of outsourcing the PCB manufacturing process, a wide variety of vulnerabilities such as tampering, recycling, and cloning etc., are being introduced to the hardware assurance community now more than ever. Therefore, from a hardware assurance perspective, it is of paramount significance to involve on-site verification under trusted conditions.

Over the years, physical inspection in the optical domain has become a popular approach within the community due to its mostly non-contact, and non-destructive nature [3]. Traditionally, a subject matter expert (SME) performs visual inspections of PCBs under certain controlled conditions. They analyze PCBs not only for defects but also for maliciously added components and trojans. But this process is time-consuming and error-prone with increasing number of PCBs. Therefore, researchers in the community have proposed different methods of image processing, computer vision, and machine learning for automatic visual inspection. However, many of these approaches rely on the existence of golden PCB, which is not always available in practice.

Application of an automatic Bill of Materials extraction method to tackle this problem was introduced in [3]. The proposed framework of materials extraction for PCB assurance (as shown in Fig. 1) involves two major steps: 1) Imaging Modality, and 2) Image Analysis. In this paper, we will focus on the image analysis step where the goal is to detect and identify the components in a PCB with high accuracy.

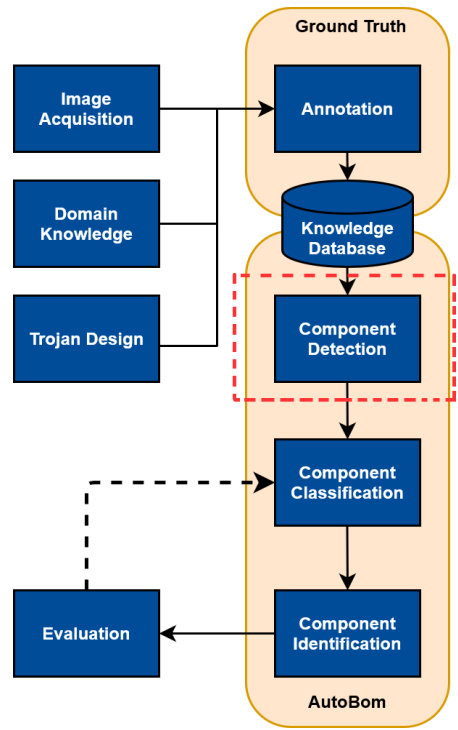

**Figure 1.** The framework for bill of materials extraction for PCB assurance as proposed in [3].

While developing an efficient system for detecting components automatically, it should be kept in mind that the method needs to be fast and highly accurate to meet the requirements of critical applications (e.g., military, biomedical applications) [3]. In addition, there are a wide variety of factors and challenges which needs special attention while developing an automated PCB assurance system. Though the majority of PCBs are monochromatic and consist mostly of standard commercial off-the-shelf components, there are an increasing variety of new and/or custom components as the state of technology is constantly advancing. Also, in the case of foreign, competitor, and/or malicious technologies, the components may be intentionally obfuscated with uncommon designs, camouflaging, and misleading or absent silkscreen labels. Moreover, factors related to image acquisition systems can also significantly affect the performance of developed component detection and identification algorithm [3]. Unfortunately, such challenges are very difficult to overcome with existing traditional computer vision (CV) and image understanding methods for component detection and identification, alone [4].

Though different machine learning (ML) and deep learning (DL) based approaches have shown significant progress in case of object detection and localization in other domains [5], it has not shown much progress in the PCB assurance domain. Though ML and DL based methods may outperform the traditional CV methods in terms of performance, there are a variety of challenges in the PCB domain that must be overcome. For example, ML and DL based methods tend to require massive amount of labelled data, which can be expensive, tedious, and time-consuming to obtain. In addition, such methods can be difficult to interpret and adapt as the state of technology is constantly advancing. To overcome these challenges, we propose the benefits of CV methods which can be used in tandem with ML and DL methods.

In this work, we explore a variety of human-interpretable image features namely, color, shape, and texture for the task of PCB component detection. Such features increase PCB assurance explainability and allow for incorporation of PCB domain knowledge. This effectively constricts the parameters of an ML or DL model and, hence, the amount of data needed to achieve a high-accuracy PCB assurance algorithm. The purpose of this paper is to facilitate collaboration between the hardware assurance, computer vision, and machine learning communities. Ours is different from the relevant works since it uses a semantic dataset instead of bounding box information for the whole components on a PCB. Unlike other approaches, ours is the first to implement computer-vision-based techniques for component detection on a PCB using semantic data. The reason behind choosing semantic information over bounding boxes is that this approach can help classify the components, track the pins, and use the pins' information to extract the corresponding netlist. This novel approach can pave the way for more future research in this direction.

The remainder of the paper is structured as follows. In Sec. II, we discussed related works and relevant existing techniques. In Sec. III, we discussed our methodology of analyzing different features using image analysis for PCB component detection. Sec. IV, V and VI contain detailed discussion on the benefits, limitations, and uses of color, shape and texture feature descriptors in hardware assurance, respectively. Sec. VII show quantitative results and analysis of the features' importances for the task of PCB component detection. Finally, in Sec. VIII, we provided an overall discussion and future research directions on top of our work followed by the concluding remarks in Sec. IX.

## 2. Related Works

In comparison to previous efforts in creating effective feature extraction methods, there has been an increased focus in deploying feature selection approaches in object identification tasks [6]. Despite the fact that numerous studies have been done to investigate the relevance of feature extraction and feature selection in machine learning-based objection detection tasks, there are few studies on applying them in the PCB assurance domain. Feature saliency, which is dependent on the size of the dataset and the difficulty of the task, determines the effectiveness of a machine learning model for object identification, segmentation, and so on. As a result, feature selection techniques are incredibly important.

According to [7], a feature can be classified as (i) very relevant, (ii) somewhat relevant but not redundant, (iii) irrelevant, or (iv) redundant. While analyzing the reasons for Convolutional Neural Networks' (CNNs) remarkable performance on complicated perceptual tasks such as object recognition, it was discovered that image texture features are more significant than object forms [8]. In [9], it was demonstrated that when color features are mixed with traditional form features, state-of-the-art results for object detection could be obtained. Due to the fact that a range of external elements can considerably impair the efficacy of a component detection method for PCB images, a variety of image attributes (e.g., color, shape, and texture) should be considered [3]. The color normalization technique [3] will help in employing an appropriate feature selection strategy for PCB component detection. In [10], a hybrid strategy for extracting color and shape features for object detection was proposed. In contrast, recent research indicates that, in addition to color and shape, texture plays a critical role in detecting objects in images [11]. As demonstrated, there are a variety of examples where CV-based features can benefit ML and DL models.

A deep learning model will perform well if sufficient data is available to train on. In the case of a limited training dataset, a deep neural network may overfit and be unable to generalize. Since a DNN has millions of parameters, each with complicated interrelationships, manually tuning the model's parameters would be incredibly challenging and computationally expensive. However, in some cases, a similar performance could be obtained by simpler methods such as basic color thresholding, which uses very little data. Certain issues can be solved using less sophisticated, and time-consuming traditional computer vision-based solutions [12]. Conventional CV-based approaches are completely transparent, but deep learning models are often criticized for being opaque and difficult to understand. This is even more challenging in the field of PCB assurance due to the variety of components, continuous changes in technology, giving rise to more added features to be considered. As DL models often suffer from a critical phenomenon known as the curse of dimensionality, dimensionality reduction is imperative to perform to limit the memory storage requirements and the computational overhead associated with data analysis. Two main components of dimensionality reduction, such as feature extraction and feature selection, offer improved learning performance, increased computing efficiency, decreased memory storage, and the development of more robust generalization models [13]. Therefore, in our work, we evaluated the importance of extracted features and proposed methods of selecting the optimal set of features strictly relevant to PCB component detection for hardware assurance.

## 3. Methodology

To prepare the data for PCB detection, we first collected PCB images and performed color correction and region windowing. Later, color, shape and texture features were extracted from each windowed region. Finally, feature selection was used to determine the importance of each feature. Importance values can be used to reduce the number of features to only the most salient ones, which can be used in tandem with ML or DL models to improve their efficiency and generalization. The process of our workfow is summarized in Fig. 2. We introduce the process in this section.

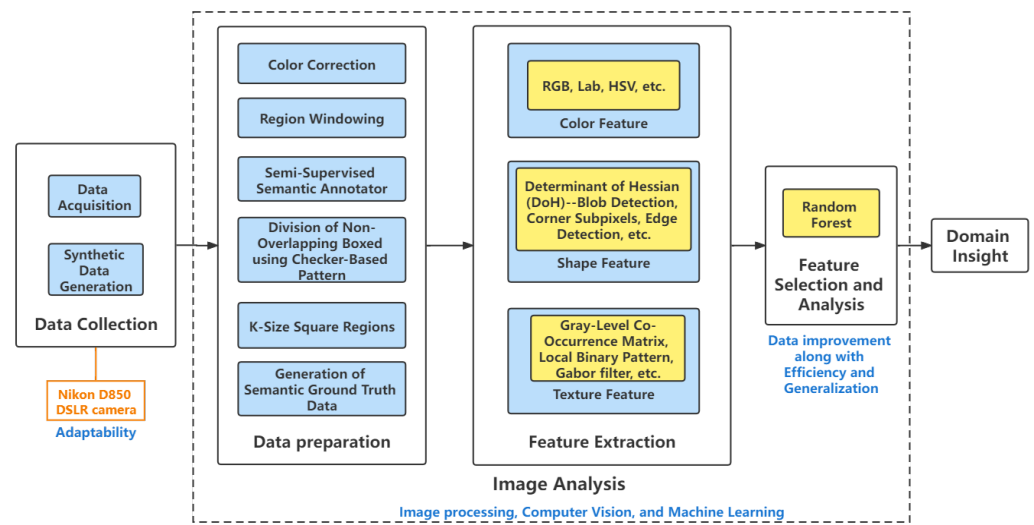

**Figure 2.** The Processing Workflow.

### 3.1. Data

In this study, we used 15 images of 10 different PCB samples, of which 5 PCBs possessed components on both the front side and the backside, so both sides were imaged. The PCBs were purchased online or disassembled from a variety of different devices such as servers, computer hard drive controllers, and audio amplifiers. Images were taken under ambient lab light with a Nikon D850 digital single-lens reflex (DSLR) camera. This camera was set to take a 2-second delay and then a 2-second exposure shot to reduce noise during image collection due to camera shake. After, color correction was used during preprocessing to effectively normalize the data in a controlled manner [3]. The dataset used here is derived from a previous study in [3].

Then, PCB components (e.g, resistors, capacitors, integrated circuits (ICs)) were annotated with semantic boundaries by human SMEs. In this study, the Semi-Supervised Semantic Annotator (S3A) was used for annotation. By dividing each image into non-overlapping boxes in a checker-board pattern, the image data was divided into square regions of different kernel sizes (ksizes). Since the ideal ksize is unknown, we performed feature extraction on each divided region for ksizes of 5, 10, 15, 20, and 25 pixels.

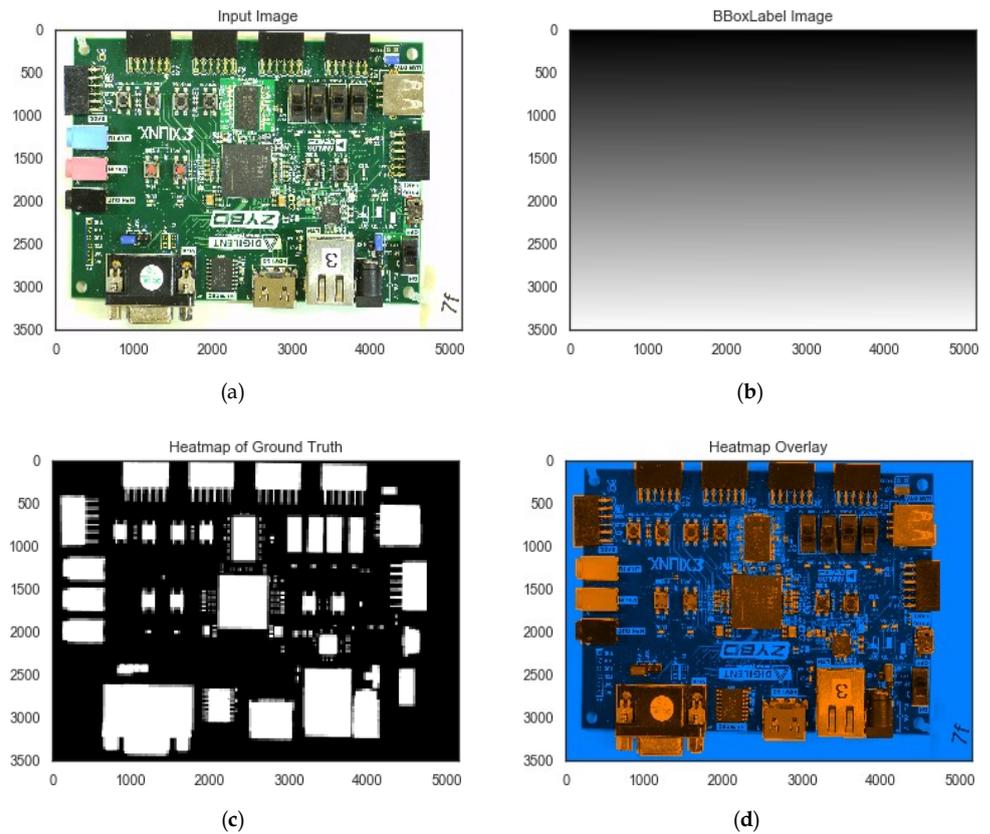

**Figure 3.** (a) The Original PCB Image; (b) The Corresponding Bbox Labels; (c) The Bbox Ground Truth Heatmap for This PCB Image; (d) The Heatmap Overlay on the PCB Image.

Since this study is concerned with PCB component detection, ground truth data was generated as follows. As feature extraction was performed on each of the non-overlapping box regions (bboxes) in the image, the ground truth needed to be in terms of the bboxes. Therefore, the bbox ground truth was generated based on the semantic ground truth data, which consisted of pixel-level labels of 1's and 0's for component and background respectively. Each bbox ground truth region was assigned values of 0-10 for the percentage of the region's pixels that correspond to a component (e.g. a region with a value of 0 did not have any component pixels in the semantic ground truth, whereas a region with a value of 10 completely consisted of component pixels in the semantic ground truth). A visual example of the bbox label mask and semantic ground truth can be seen in Fig. 3.

*3.2. Feature Extraction*

Feature extraction of image is the key to the image recognition process. Selecting appropriate image features for different components can improve the efficiency of components recognition. In our study, we have extracted features of PCB board images from three aspects: color feature, shape feature, and texture feature.

Color is intuitively an important feature for detecting components, as many components are distinct in color from the PCB board. For example, a black SMD voltage resistor is distinct in color from a monochromatic green PCB, but not a black board. Color features can show good stability in different shapes and placement directions. Here, 13 different methods to extract color features were implemented, of which 3 methods will be discussed in detail in Section IV.

In addition, shape features are also intuitively an important feature for detecting PCB components, as many components consist of regular shapes. For example, many SMD resistors are rectangular, while many vias on the PCB appear circular from above. Shapes are crucial for humans to distinguish different objects, so they are regarded as very critical and important in computer vision and pattern recognition. There are many ways to express the shape of an object in a computer: local/boundary shape features and global shape features. Local shape features require prior segmentation, such as area, perimeter, etc.. Global shape features do not require segmentation and can be computed on the entire image, such as edge, corner, and blob detection. Since this study concerns PCB component detection, i.e. prior segmentation is not known, we used global shape features. Three feature extraction methods will be discussed in Section V.

Finally, texture features are also intuitively beneficial for detecting PCB components, as the components are often composed of different materials which can possess different patterns when imaged. For example, plastic packaging materials on an IC tend to appear rougher and less reflective than the head of some ceramic capacitors. The texture feature is an image feature that reflects the spatial distribution of pixels, and it is usually characterized by local irregularities and macroscopic regularities [14]. Different image grayscale pixel arrangements will produce different texture features to different components for distinction. Statistical methods and signal processing methods are the two main texture feature methods. Here, 9 texture feature extraction approaches were implemented, 3 of which will be discussed in Section VI.

### 3.3. Feature Selection and Analysis

Feature selection is used to extract the optimal subsets for PCB component detection [15]. It can effectively eliminate irrelevant features, reduce data dimensions, and improve the accuracy and efficiency of classification models. In this step, we use the feature selection algorithm to rank the importance of the 1200 features extracted in the previous step.

Commonly-used feature selection algorithms are broadly categorized into filters, wrappers, and embedding methods [16]. At present, the feature selection mechanisms are mainly based on Information Theory, Neural Networks, Support Vector Machine (SVM) [17] [18], etc.

Breiman proposed the Random Forest (RF) algorithm in 2001. This method operated by building a large number of decision trees and belongs to an ensemble algorithm [19]. Random forests can give the importance of features by calculating the average impurity reduction based on all decision trees in the forest. It won't need to make any assumptions about the linear separability of data [20].

In the process of splitting decision tree nodes in a random forest, we define the Gini importance [21] $i$ as:

$$i = 1 - \sum_j p^2 \quad (1)$$

The proportion of samples marked j in the node is $p(j)$. After the split, the importance of Gini is:

$$\Delta i = i_{parent} - (p_{left} \cdot i_{left} + p_{right} \cdot i_{right}) \quad (2)$$

Among them, the sample proportions of the left child node and the right child node are $p_{left}$ and $p_{right}$. $i_{parent}$, $i_{left}$, and $i_{left}$ respectively represent the Gini importance of it parent node, the left child node, and the left child node. For any feature $X_i$, the decreasing sum of impurities in all decision trees is the Gini importance of $X_i$:

$$\alpha \Delta I = \sum_k \Delta i_k \quad (3)$$

Based on this equation, this value indicates the importance of each feature.

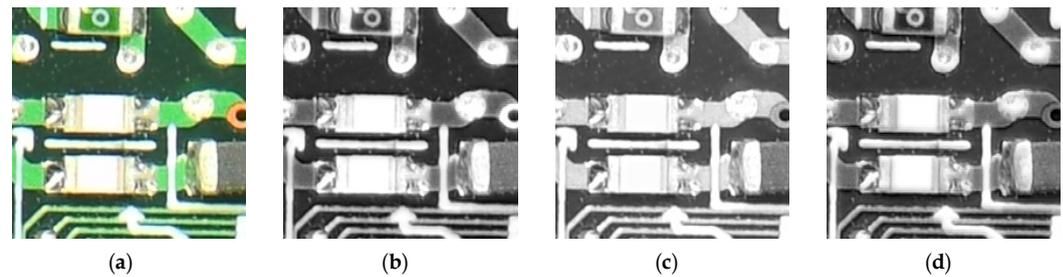

**Figure 4.** (a) The Original PCB Image; (b) R Channel of the Image; (c) G Channel of the Image; (d) B Channel of The Image.

**4. Color Features**

To better express the color feature of PCBs, we used 13 methods, including RGB, RGB_CIE, HSV, HLS, LAB, LUV, YCrCb, YDbDr, YPbPr, XYZ, YIQ, YUV, HED. In this section, we discuss 3 kinds of color features, which are RGB, HSV and Lab color feature. These three color features play an important role in image feature extraction.

*4.1. RGB*

The RGB color space is the most common representation for the color of pixels. The RGB color space uses the superposition of three primary colors in physics to form the principle of producing various colors. In RGB color space, the attributes of the three components of R, G, and B are independent, respectively representing the color of red, green and blue. The gray-scale images of three channels in RGB color space and the original image are shown in Fig. 4.

4.1.1. Benefits

RGB is the most commonly used and most basic way of expressing color characteristics, and the expression of different components on the PCB is more intuitive and easy to understand.

4.1.2. Limitations

RGB color space is the most common hardware-oriented model, which is usually used in imaging and display systems, and is rarely used in image processing and feature extraction [22]. The color components of RGB space may be affected by light and the environment. The three color components are highly correlated, The brightness will change with the transformation of the three components of R, G, and B. Moreover, when one of the color components changes, it will also affect the other two color components to a certain extent. Two colors with the same chromaticity might be mistaken for each other when their intensities change even by the slightest amount. The challenging cases include distinguishing surface-mount resistors and inductors on a PCB.

*4.2. HSV*

A. R. Smith created the HSV color space in 1978 based on the intuitive characteristics of colors, and it is also called also called Hexcone Model [23]. This HSV model respectively represents hue (H), saturation (S), and value (V). To express color on the image, the Hue refers to an image's relative lightness and darkness. Saturation shows the proximity between color to the spectral color. The value indicates how bright the color is. HSL is similar to HSV in the color space and they both are related to the concept of the human visual system [24]. However, they are slightly different in conceptual expression [25]. Fig. 5 shows the PCB image represented on the HSV color space.

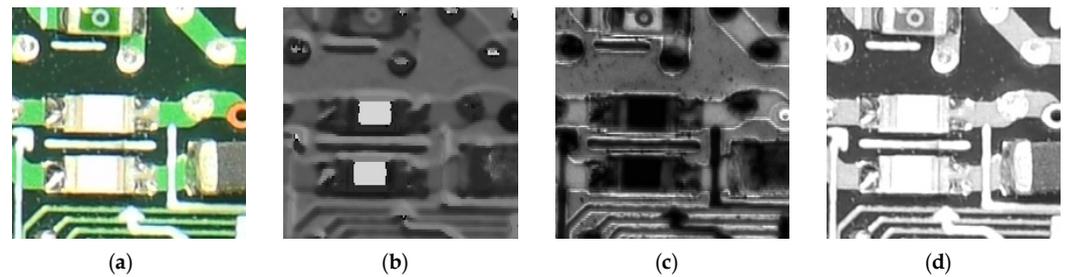

**Figure 5.** (a) The Original PCB Image; (b) H Channel of the Image; (c) S Channel of the Image; (d) V Channel of The Image.

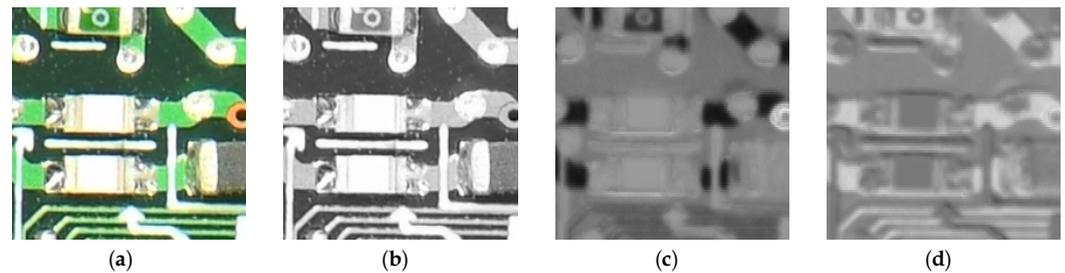

**Figure 6.** (a) The Original PCB Image; (b) L Channel of the Image; (c) A Channel of the Image; (d) B Channel of The Image.

4.2.1. Benefits

Each attribute of HSV directly corresponds to the basic color concept, which makes it conceptually simple and easy to understand. HSV can eliminate the influence of intensity components from the color information carried in color images. When the external illumination environment fluctuates slightly (as is frequently the case when optical images of PCBs are acquired), Hue values vary less than RGB values. For instance, two colors of red may have comparable Hue values yet vastly dissimilar RGB values. Thus, when differentiating identical components on a PCB under varying lighting conditions, the HSV color space may produce a more intuitive result.

4.2.2. Limitations

HSV is not suitable for use in illumination models. Many luminous mixed operations and luminous intensity operations cannot be implemented directly using HSV. A major disadvantage of the HSV spaces is that white, black, and gray do not have a distinct chromaticity; consequently, these colors are treated as singularities, making them difficult to detect. So the components, e.g., resistors, inductors, diodes and ICs with the surfaces of these colors will be difficult to deal with using this color space.

*4.3. Lab*

The International Eclairage Committee (CIE) developed the Lab color model in 1931, which is improved in 1976 and named as a international standard color mode for color measurement. The L component represents the lightness of the pixel. a and b respectively represent the color range from from red to green, and from yellow to blue. Fig. 6 indicates the PCB image represented in Lab color space.

4.3.1. Benefits

This color space can directly compare and analyze different colors by using the geometric distance in the color space. Certain kinds of components in PCB don't have visible differences to the naked eye, like the capacitors and the resistors, or different kinds of chips. Lab feature has a wide color gamut so it can be effectively and conveniently used to measure slight color differences, such as surface-mount resistors and inductors.

### 4.3.2. Limitations

The creation of Lab space is relatively complicated, and the Lab color space generated is not as natural and understandable for humans as RGB or other perceptual color spaces[26]. This color space suffers from the same singularity issue as discussed in the limitations of the HSV color space.

In addition to the above-discussed color features (RGB, HSV, and Lab), the shape and texture features will help to detect the components of PCBs more efficiently that are discussed in the Feature Extraction section. We will converse about few significant shape features in-depth in the following section.

## 5. Shape Features

In our study, there are 11 types of shape features: Histogram of Gradients (HOG) [27], Scale Invariant Feature Transform (SIFT) [28], Oriented FAST and Rotated BRIEF (ORB) [29], Hough Line Transform, Hough Circle Transform [30], Determinant of Hessian (DoH) - Blob Detection [31], Fourier Transform, Connected Components, Corner Subpixels, Local Peak Maxima, and Edge Detection. In this section, we will discuss the following three shape features: Determinant of Hessian (DoH) - Blob Detection, Corner Subpixels Detection, and Edge Detection features. These three shape features are significant in the extraction of image features.

### 5.1. Determinant of Hessian (DoH) - Blob Detection

The blob objects are generally bright on dark regions or dark on bright regions on an image and can be extracted using three algorithms. One algorithm computes Laplacian of Gaussian (LoG) [32] with consecutively increasing standard deviation and piles them as a cubic structure. Local maxima of the cube are considered as blobs. This procedure executes very slow on extracting larger blobs due to a high number of convolutions and only bright objects are detected in dark regions. An alternate method to LoG, Difference of Gaussian (DoG) [32], works on approximation by blurring the image during the convolution, so the difference between two consecutive blurred images are piled up as a cubic structure. Unfortunately, this algorithm fails to detect larger blobs efficiently. The third algorithm, Determinant of Hessian (DoH), detects blobs using the DoH matrix by computing maxima. This method uses box filters instead of convolutions which removes the dependency of blob sizes in execution. Both bright and dark blobs are detected through this procedure.

#### 5.1.1. Hyperparameters

We have set the hyperparameters by tuning them to get appropriate results as shown in Fig. 7. The parameters are: minimum and maximum $\sigma$ or standard deviation for Gaussian kernel, the threshold which is lower bound for scale-space maxima, overlap value that determines area limit of two blobs overlapped, and log scale set to default value i.e., False.

#### 5.1.2. Benefits

Any circular and non-circular curvy components such as oscillators, transistors, and a number of diodes are easily detected using DoH blob features from PCB images.

#### 5.1.3. Limitations

This algorithm will not be able to detect small blobs accurately such as vias on a PCB.

### 5.2. Corner Subpixels

We can represent corners as the pixel points with huge intensity variation from all directions around the pixel [33]. According to Harris & Stephens [34], the corners are identified based on the difference in the intensity for a displacement of $(u, v)$ in all directions as shown below:

$$E(u, v) = \Sigma_{x,y} w(x, y)[I(x + u, y + v) - I(x, y)]^2 \quad (4)$$

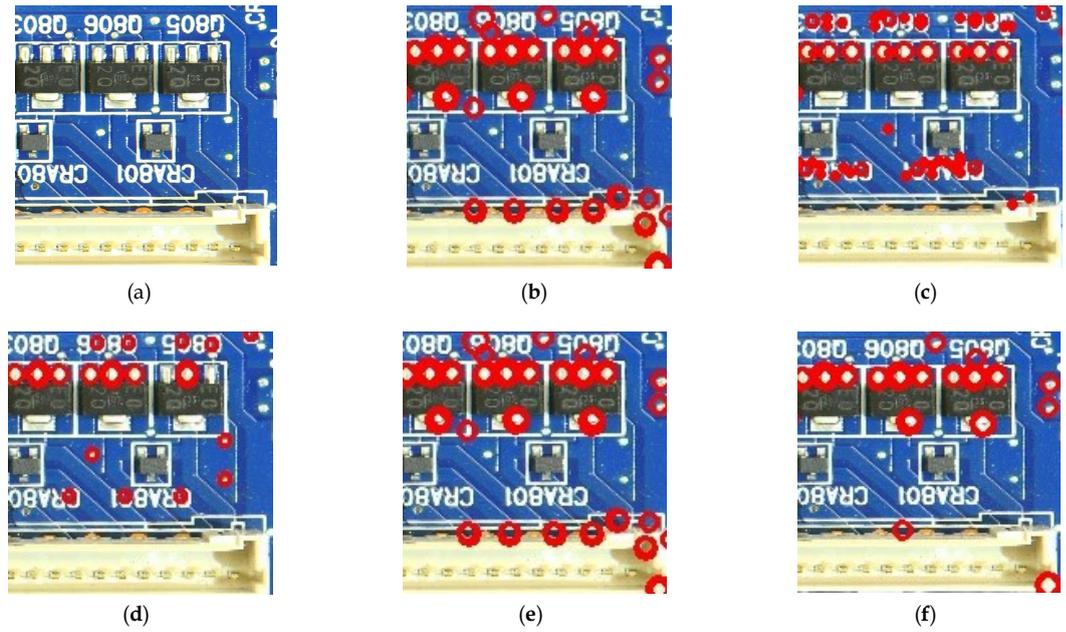

**Figure 7.** Determinant of Hessian - Blobs feature images with different label mask k-size (a) Original Image Patch, and (b-f) Respective experimental result from 25 to 5 image mask size.

The window function $w(x, y)$ could be either a gaussian window or rectangular window used to add weights to pixels. To detect a corner, we need to maximize the $E(u, v)$ function. In the process of maximizing, we can apply Taylor expansion to the above equation that produces the below equation:

$$E(u, v) \approx \begin{bmatrix} u & v \end{bmatrix} M \begin{bmatrix} u \\ v \end{bmatrix} \quad (5)$$

$$\text{where } M = \sum_{x,y} w(x, y) \begin{bmatrix} I_x I_x & I_x I_y \\ I_x I_y & I_y I_y \end{bmatrix}$$

In the above equation, $I_x$ and $I_y$ are derivatives of the image in x and y directions, respectively. We can determine whether a point is a corner or not by computing the scoring function $R$ as below. If the value of $R$ is large, it implies $\lambda_1$ and $\lambda_2$ are large and equivalent, then it is a corner.

$$R = det(M) - k(trace(M))^2 \quad (6)$$

$$\text{where } det(M) = \lambda_1 \lambda_2, \; trace(M) = \lambda_1 + \lambda_2$$

In contrast to Harris corner algorithm, Shi-Tomasi [35] proposed a different approach of calculating the scoring function ($R$) as below which produces better results compared to previous algorithm.

$$R = min(\lambda_1, \lambda_2) \quad (7)$$

If $\lambda_1$ and $\lambda_2$ are greater than the minimum threshold R, it can be considered as a corner. Later, we apply refinement to the detected corners using the corner subpixel algorithm.

5.2.1. Hyperparameters

In this experiment, we have two primary functions: Shi-Tomasi's *GoodFeaturesToTrack* and *CornerSubpix*. The first function takes the number of corners, quality level, minimum distance, block size, and gradient size as the hyper-parameters that vary accordingly to the label mask's k-size. Similarly, the second function takes three parameters window size, zero zone, and criteria that are constant irrespective of the label mask's k-size. The results are shown in Fig. 8.

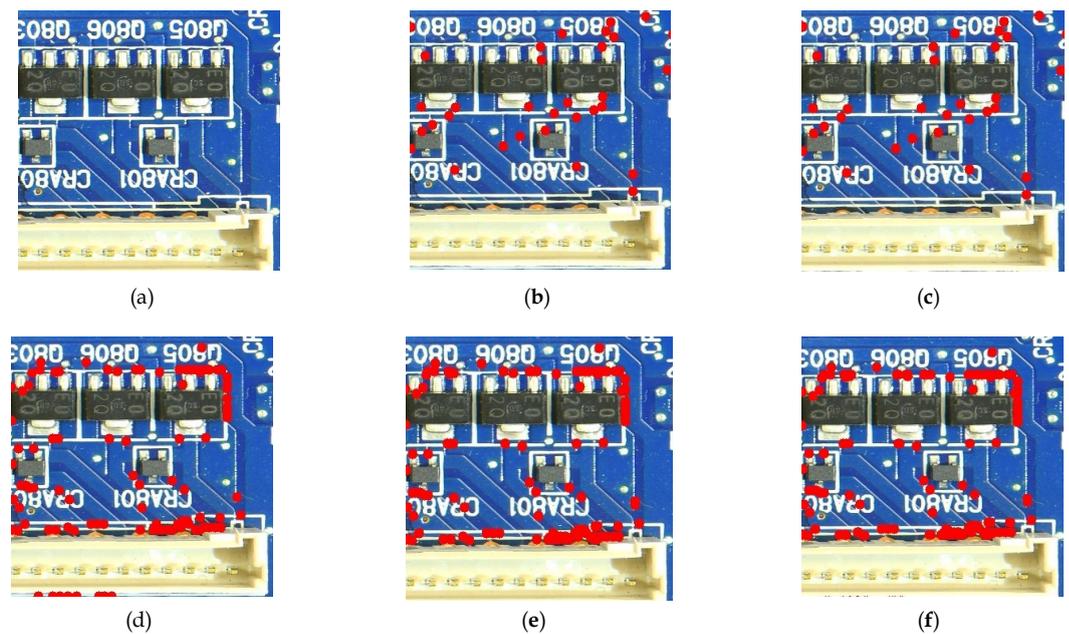

**Figure 8.** Corner Subpixel feature images with different label mask k-size (a) Original Image Patch, and (b-f) Respective experimental results from 25 to 5 image mask size.

5.2.2. Benefits

Similar to Connected Components features, these features are scale, rotation, and illumination invariant. Using these corner features, we can easily identify the location of the component on the PCB.

5.2.3. Limitations

These features consume more memory space to process the algorithm due to redundancies and are not very robust in a complex image, for example, the image of a PCB with a high density of components or reference designators.

5.3. *Edge Detection*

Edge detection is the process of finding consecutive points of a sudden change in brightness that forms an edge in an image. The major steps involved in edge feature detection are gray-scaling, bilateral filtering for noise removal, edge detection using Canny algorithm [36], identifying contours around the detected edges, computing statistics such as the number of contours, maximum contour area, etc. as features. The Canny algorithm finds the gradients of the blurred image and utilizes the non-maximum suppression along with the hysteresis for removing spurious edges and weak edges from the detections, respectively. The canny algorithm has been used due to its adaptability to variations in the images.

5.3.1. Hyperparameters

There are three different hyperparameters for each step. The diameter is set to 7, and sigma color and sigma space are set to 50 for the bilateral filtering. For the Canny edge algorithm, the lower threshold is set to (mean of gray intensities – 25% of the mean), the upper threshold is set to (mean of gray intensities + 25% of the mean), and *L2 Gradient* to false. Contour finding expects two significant parameters i.e., retrieval mode is set to *RETR_EXTERNAL* and approximation method to *CHAIN_APPROX_NONE*.

### 5.3.2. Benefits

This edge feature detects both smaller and larger components on PCBs such as ICs, diodes, transistors, etc. The Canny edge detector aids in the detection of lines, which is advantageous for trace analysis in PCBs.

### 5.3.3. Limitations

These edge features could get biased towards horizontal and vertical edges in the images. There is a likelihood of wrong approximations of symmetry on rotations that are common in most PCBs, especially for ICs with square shapes.

Although the shape features play a significant role in image feature extraction, they are alone not sufficient enough to uniquely identify and detect the components on the PCB. As we have mentioned that texture indicates the spatial distribution of pixels, the texture features along with shape and color can improve the accuracy of detecting the PCB components. In the next section, we will explore the texture features in detail.

## 6. Texture Features

For texture features, we have used 9 kinds of texture feature extraction methods: Gabor filter, Gray-level co-occurrence matrix (GLCM), Local binary pattern (LBP), ray-level run length matrix (GLRLM), Tamura, Law's Texture Energy Measures (LTEM), Gray-level difference statistics, Autocorrelation function, and Segmentation-based fractal texture analysis (SFTA). Among these 9 methods, GLCM, LBP, and Gabor filter have a more obvious effect of distinguishing components, and they are also the most common feature extraction methods for image texture features. In this section, we will introduce and analyze these three methods.

### 6.1. Gabor filter

To better describing the texture information of the image, the first method we choose is Gabor filter. It's usually used in signal processing. To describe the local frequency information of the image signal, the Gabor kernel adds a window function to the signal in the frequency domain [37]. Gabor filter kernel is similar to the receptive field of vertebrate visual cortex [38], which provides a good result for texture representation and discrimination [39].

#### 6.1.1. Hyperparameters

$\lambda$ represents the wavelength of the filter function. The longer the wavelength, the greater the interval between black and white stripes in the Gabor kernel image. $\theta$ represents the tilt angle of the kernel function image, which can be used for effective feature extraction for textures in different directions. $\psi$ determines the phase shift. When $\psi$ is 0, the kernel center is a white stripe; when $\psi$ is 180, the kernel center is a black stripe. $\sigma$ is the standard deviation of the Gaussian function, which reflects the effective size of the kernel. $\gamma$ is the spatial aspect ratio, which determines the ellipticity of the filter kernel function[40].

After testing, when $\lambda = 14$, $\psi = 0$, $\sigma = 5$, $\gamma = 1$, the texture feature extraction effect is the best. Considering that the components on the PCB may have different placement directions, we set 6 values for $\theta$ : 0°, 30°, 60°, 90°, 120°, 150°. The output images after applying Gabor filtering are shown in Fig. 9.

#### 6.1.2. Benefits

For the kernels, we have chosen 6 directions, which can basically cover the direction of all components on the PCB board, so the Gabor filter has the characteristic of rotation invariance. For a certain degree of image rotation and distortion, Gabor filter can still

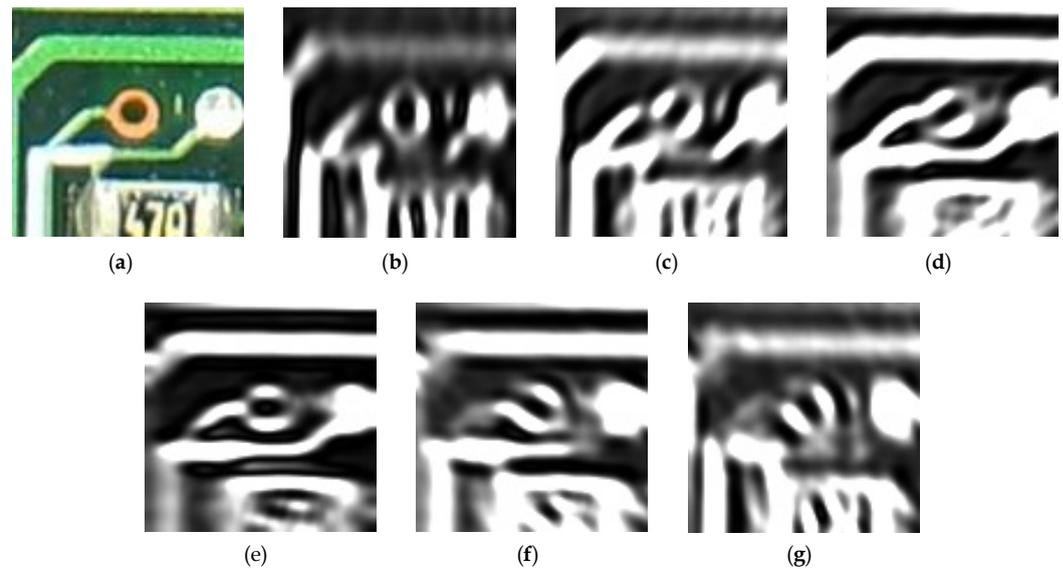

**Figure 9.** (a) A part of the Original PCB Image; (b) Filtered Image when *θ* = 0°; (c) Filtered Image when *θ* = 30°; (d) Filtered Image when *θ* = 60°; (e) Filtered Image when *θ* = 90°; (f) Filtered Image when *θ* = 120°; (g) Filtered Image when *θ* = 150°

provide better results. In addition, Gabor filter is insensitive to light changes. If the light at each position on the PCB is not exactly the same, it is able to provide a good adaptability.

6.1.3. Limitations

The limitation of the Gabor filter is that it is non-orthogonal, which will result in different proportions of redundant features [41]. In addition, due to the high frequency response at the image edge, a "ring" effect may occur [42]. This has potential to create problems while detecting vias on a PCB.

*6.2. Gray-level co-occurrence matrix*

Gray-level co-occurrence matrix (GLCM) was proposed by Haralick et al. in 1973, which refers to a simple way to describe textures by studying the distribution of different gray levels corresponding to different textures in space [43]. That is, the spatial correlation in grayscale images [44].

6.2.1. Hyperparameters

We select four angle directions of 0°, 45°, 90°, 135° to calculate GLCM. With having a small ksize, the best step size should be set to one pixel, otherwise the output images will become blurry. We also have changed the 8-bit pixel to 4-bit (which means the gray-level is equal to 16) to improve the computational efficiency. The output images when the angle is 0° is shown in Fig. 10.

6.2.2. Benefits

GLCM has strong adaptability and robustness. Its features can be produced for a single orientation or for a group of orientations, making GLCM direction independent, which may effectively cover the orientation of all components on the PCB.

6.2.3. Limitations

As a statistical method for texture feature extraction, GLCM has less correlation with human visual models and lacks the use of global information. With high computational complexity, GLCM has long execution time. This technique is not suitable for distinguish-ing between different text fonts that is why this feature faces difficulty when detecting component markings and reference designators on a PCB.

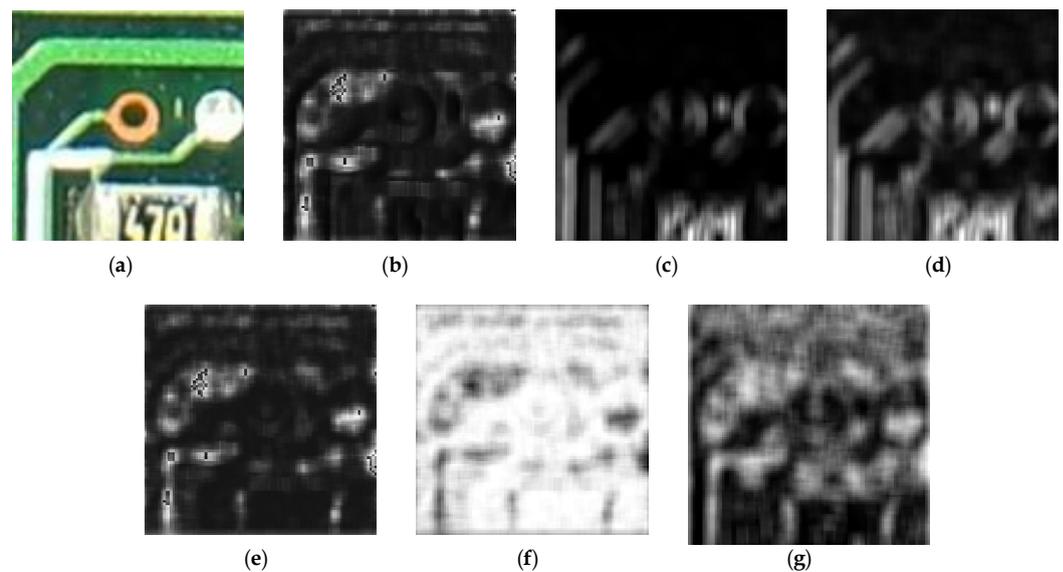

**Figure 10.** (a) A Part of the Original PCB Image; (b) ASM Image when $\theta = 0°$, (c) Contrast Image when $\theta = 0°$; (d) Dissimilarity Image when $\theta = 0°$; (e) Energy Image when $\theta = 0°$, (f) Entropy Image when $\theta = 0°$; (g) Homogeneity Image when $\theta = 0°$.

### 6.3. Local binary pattern

LBP reflects the texture changes around the image pixels. We define a $3 \times 3$ window as the LBP operator. For the center pixel in the window, compare the pixel values with its neighboring 8 pixels. Mark the surrounding positions as 0 or 1 according to the compared result. Greater than the central pixel value is recorded as 1, otherwise it is recorded as 0. In this case, we get an 8-bit binary number, and then we convert it to a decimal number. Use the obtained decimal value to reflect the texture information of the $3 \times 3$ area, which is also the LBP value of the center pixel of the window [45].

The LBP at this time is available to represent texture feature, but not rotation-invariant. As the image rotates, the pixel position changes, and the LBP value will change accordingly, and the given feature values will be very different. Maenpaa et al. proposed a method to achieve Rotated Local Binary Pattern (LBP) operation: continuously rotate the circular field to obtain all possible initial defined LBP values, and then take the minimum value as the value of the field [46] [47].

Next, in order to further improve statistical capabilities, it is necessary to solve the problem of excessive binary patterns. Ojala et al. came up with a unified mode adaptation to reduce the dimensionality of the LBP operator mode type, which is called the unified local binary mode (ULBP). When the cyclic binary number corresponding to a certain LBP changes from 0 to 1 or from 1 to 0 at most two transitions, the binary corresponding to the LBP is called a uniform pattern class.

In here, we have combined the RLBP and the ULBP to form the most powerful rotated uniform LBP feature.

#### 6.3.1. Hyperparameters

In the calculation process of this algorithm, a sliding window is needed. If the window is too small, there will be mis-segmentation within the same texture; while when the analysis window is too big, there will be mis-segmentation in the texture boundary area. Since the region size is relatively small and the window size is $3 \times 3$, the testing result shows that the texture effect has the best form. We can see the output image in Fig. 11.

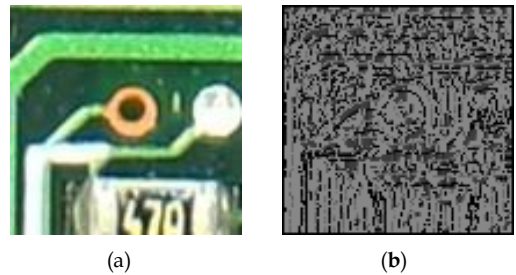

(a)  (b)

**Figure 11.** (a) A Part of the Original PCB Image; (b) The Output Image after RLBP Operators and ULBP Operators.

6.3.2. Benefits

The RLBP and the ULBP are combined to retain the most effective feature value, and this method has rotation invariance, gray scale invariance, and invariance to monotonic illumination changes [43]. This helps with the same type of components with different orientations on a PCB in variant lighting conditions.

6.3.3. Limitations

The calculation time is positively correlated with to the number of pixels in the image. As the image grows larger, it will take longer to perform calculations [48] Moreover, noise, blurring, and any other disturbing effect, all have noticeable impacts on the method performance. So, in case of blurry markings or having reference designators and incorporated noise due to image acquisition, LBP texture feature might not be suitable for PCB assurance.

In the next section, we will compare the different results from the feature extraction experiment to recognize the importance of few features based on their impact on PCB component detection.

## 7. Results

Among the employed 34 feature extraction methods, 13 of them are color, 12 for shape, and 9 are for texture. Overall, we have used these 34 feature extraction methods to obtain a total of 1200 PCB board image features. And After feature selection, we got the output data of the importance of different features with different ksizes. Bounding boxes with different sizes usually have a certain impact on the feature extraction results.

The box plot is used to depict the center and spread of the data distribution, so here we use the box plot to represent the feature data distribution under different ksizes, as shown in Fig. 12.

In the box plot, each group of ksize data is composed of the sum of corresponding features in 15 PCB board images. The two ends of the box are the upper and lower quartiles, which are the median of numbers larger than the overall median, and the median of numbers smaller than the overall median. The height of the box, which is the distance between first quartile and the third quartile, reflects the degree of data fluctuation. So the distribution of the characteristic data of ksize15 is more scattered, while for ksize5 is more concentrated. Although the overall eigenvalues of ksize5 are close to 0, so it is not a proper choice.

According to the higher median and shorter distance shown in the boxplot of Figure 12, ksize of 25 has the highest significance among the 5 different ksizes. This result is explained based on the fact that the smaller ksizes are very unfavorable for extraction of the shape and texture features. For example, it is difficult for a small region in the shape feature to reflect the lines and corners, while it is difficult to reflect the repeatability among pixels in the texture feature.

The distribution of the importance of the color feature values is generally higher than shape and texture features. Since the regional features of the image are related to the entire

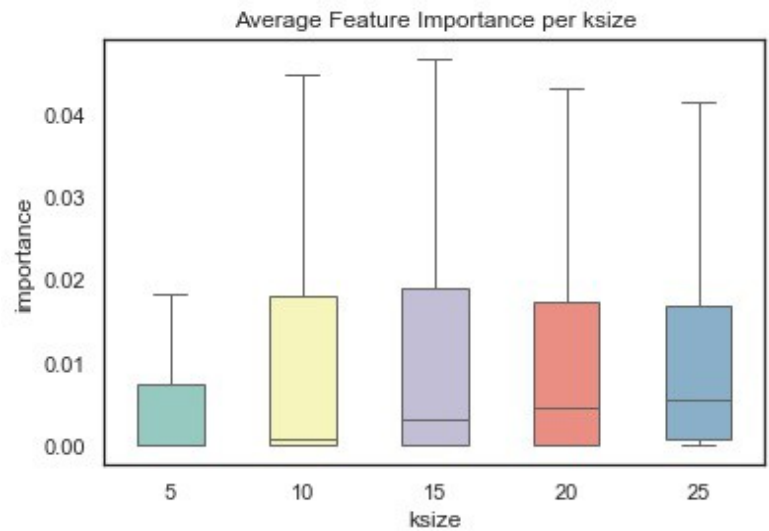

**Figure 12.** The Boxplot for Different Ksizes. It indicate that ksize 25 has the higher median and shorter distance, so ksize 25 is the best among the 5 different ksizes.

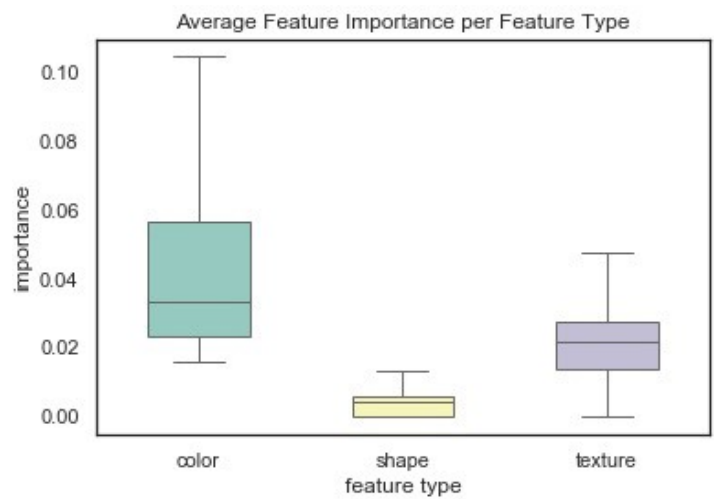

**Figure 13.** The Boxplot for Different Feature Type in Different Images. The distribution of the color feature in the box plot shows that it is the most effective feature among the three types of features

shape area, the shape features in a single area of the PCB board image after bounding box segmentation are not ideal. The distribution of feature types are displayed in Fig. 13.

Through sorting the features base on their importance, we have selected the five most important features that are "HLS_2_mean", "LAB_1_med","LAB_1_mean", "HED_1_med", "HED_1_mean". Note: in this study, color features were identified by their color space, channel, and the function used to compute a single feature value from each region. For example, the most important feature "HLS_2_mean" indicates the mean of the second channel of each region when the image was converted to HLS color space was considered most important. The distribution of these feature values in 15 images are shown in Fig. 14. Those five significant features belong to the color type.

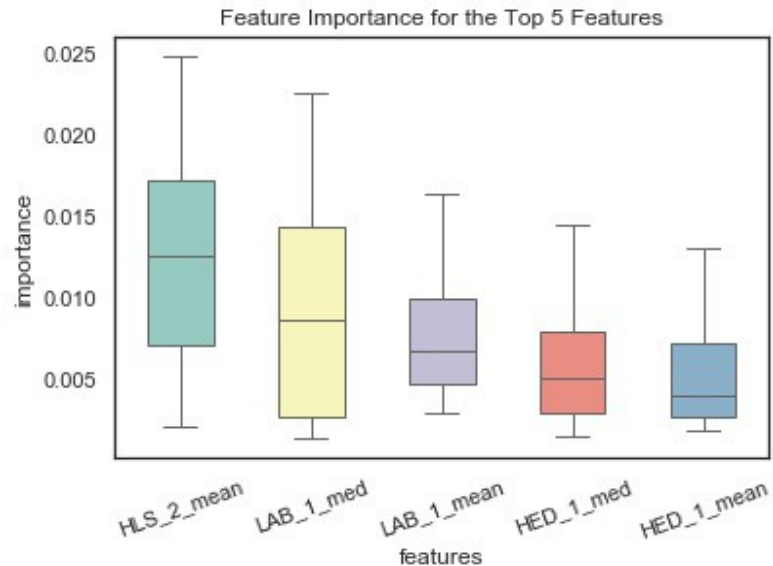

**Figure 14.** The Boxplot for the Five Most Important Feature Types in Different Images.The top five important features all come from color features, this also shows that the color feature is the most important among the three types of features.

## 8. Discussion

Overall, the results of this study demonstrate color features are completely informative and useful PCB component detection. As shown in Section VII, color features for most cases demonstrated higher levels of importance in PCB component detection than shape and texture features. In other words, color features are generally more informative for the task of PCB component detection. The higher strength of color features is due to monochrome property (a single base color) of PCB boards and the face that they process a distinct color form the components. Therefore, we consider color features as great candidates for modeling the PCB boards with satisfactory performance. With respect to execution time, color features tend to be the fastest to get extracted because no parameter tuning is necessary and many of the algorithm operations are vectorised and this trait of color features represents a promising prospect for real-time PCB assurance. Table. 1 represents a comparative analysis between existing works and our study. It is evident from the table that our work demonstrates a novel approach to investigate the impact of different features in recognizing certain components on the PCB using semantic image dataset.

## 9. Conclusions

In this study, we have found color features overall are faster to extract and more accurate for PCB component detection than shape and texture features. It is important to note that this is a controlled experiment, as all PCB images used in this study were obtained using similar lighting conditions. Since color features are sensitive to such conditions (e.g.

**Table 1.** Comparative analysis between existing works and our study

| Papers | Dataset | Use cases | Method | Result |
|---|---|---|---|---|
| [49] | CAD files of the PCB and bare PCB image datasets | PCB inspection | LIF (Learning Inspection Features) and OLI (On-line Inspection) | Detection accuracy exceeded 97%. |
| [50] | Bounding box PCB image datasets | Detecting specific PCBs and recognizing mainboards | ORB features and Random Forest | The PCB recognition accuracy is 98.6% and the classification accuracy is 83%. |
| [51] | Bounding box PCB image datasets | Component analysis, IC detection and localization | YOLO, Faster-RCNN, Retinanet-50 | The mean average precision of these 3 techniques are: 0.698, 0.783 and 0.833. |
| [52] | Semantic PCB image datasets | PCB element detection | SSD neural network | The mean average precision of normal, enhanced, and ideal images are 0.9209, 0.9272, and 0.9510. |
| [53] | Bare PCB image datasets | Defect detection and classification | Image processing and flood fill operation | Classified up to 7 defects and the defects are identified successfully. |
| Our work | Semantic PCB image dataset | Analyzing a variety of common computer vision-based features for the task of PCB component detection | 34 feature extraction methods for color, shape, and texture; Random Forest | For most of the cases, color features demonstrated higher levels of importance in PCB component detection than shape and texture features. |

a brown capacitor imaged under dim light could appear similar to a black resistor), a color normalization technique and prepossessing algorithms should be utilized prior to feature extraction to ensure better generalizability.

Though shape and texture features were generally slower to extract and less accurate for PCB component detection than color features, they could still be very useful for PCB assurance. For example, after the components have been detected, local shape features such as size, aspect ratio, and shape complexity could be helpful for classifying the different PCB components, as many component types have a distinct footprint (e.g. 3-prong transistors vs. 8-pin DIP ICs). In addition, texture features such as regional smoothness and variance could be helpful for detecting visual defects such as scratches, spurious copper, and mousebites. Since both shape and texture features are sensitive to lighting conditions as well as imaging resolution, both color and shape normalization techniques and preprocessing algorithms should be utilized prior to feature extraction to ensure better generalizability.

Our work and data will be applied to do feature selection for PCB component detection using semantic data. In future works, the tradeoffs between using more and diverse data and as well as transferring knowledge in hardware assurance applications should be more comprehensively explored. Employing large high number of varied data samples would be effective for common cases for which we have lots of examples (such as off-the-shelf components), but such data is time-consuming to collect and can be very difficult to acquire on competitor or foreign custom components, legacy devices, and malicious modifications. On the other hand, leveraging strong prior knowledge prior knowledge would increase the explainability of the system and reduce the amount of data needed for difficult-to-obtain cases. However, domain knowledge can be difficult to translate into an algorithm for certain data types. Since the hardware assurance domain is a complex and constantly evolving field, there is a great demand for both approaches. All in all, this work can be considered as a new direction and motivation for the artificial intelligence and computer vision communities to get involved in hardware security studies and vice versa.

**Acknowledgments:** Olivia Paradis, for collecting the data, conceptualizing the experiment, and writing code.

**Conflicts of Interest:** The authors declare no conflict of interest.